\documentclass[a4paper,fleqn]{cas-sc}

\usepackage[authoryear,longnamesfirst]{natbib}
\usepackage {enumitem}
\usepackage {enumerate}
\usepackage{graphicx}
\usepackage{lineno}
\usepackage{perpage} 
\MakePerPage{footnote}

\def\tsc#1{\csdef{#1}{\textsc{\lowercase{#1}}\xspace}}
\tsc{WGM}
\tsc{QE}
\tsc{EP}
\tsc{PMS}
\tsc{BEC}
\tsc{DE}

\begin{document}
\let\WriteBookmarks\relax
\def\floatpagepagefraction{1}  
\def\textpagefraction{.001}

\shorttitle{Vocabulary expansion of multilingual language models}

\shortauthors{Jianyu Zheng}

\title [mode = title]{Effective vocabulary expansion of multilingual language models for extremely low-resource languages}                      


%
\author[1,2]{Jianyu Zheng}[style=chinese,
                         orcid=0009-0007-1810-5010]

\cormark[1]


\ead{zheng_jianyu@126.com}


\affiliation[1]{organization={School of Foreign Languages, University of Electronic Science and Technology of China},
    addressline={Chengdu}, 
    city={Sichuan Province},
    postcode={611731 CN}, 
    country={China}}
\affiliation[2]{organization={School of Computer Science and Engineering, University of Electronic Science and Technology of China},
    addressline={Chengdu}, 
    city={Sichuan Province},
    postcode={611731 CN}, 
    country={China}}

\cortext[cor1]{Corresponding author}

\begin{abstract}
Multilingual pre-trained language models(mPLMs) offer significant benefits for many low-resource languages. To further expand the range of languages these models can support, many works focus on continued pre-training of these models. However, few works address how to extend mPLMs to low-resource languages that were previously unsupported. To tackle this issue, we expand the model's vocabulary using a target language corpus. We then screen out a subset from the model's original vocabulary, which is biased towards representing the source language(e.g. English), and utilize bilingual dictionaries to initialize the representations of the expanded vocabulary. Subsequently, we continue to pre-train the mPLMs using the target language corpus, based on the representations of these expanded vocabulary. Experimental results show that our proposed method outperforms the baseline, which uses randomly initialized expanded vocabulary for continued pre-training, in POS tagging and NER tasks, achieving improvements by 0.54\% and 2.60\%, respectively. Furthermore, our method demonstrates high robustness in selecting the training corpora, and the models' performance on the source language does not degrade after continued pre-training.
\end{abstract}


\begin{keywords}
multilingual pre-trained language model \sep continued pre-training \sep expanded vocabulary \sep bilingual dictionary
\end{keywords}

\maketitle

\section{Introduction}
The emergence of multilingual pre-trained language models (mPLMs) brings significant benefits to a wider range of languages\citep{doddapaneni2025primer}. These models are trained on large-scale multilingual corpora through tasks such as Masked Language Modeling (MLM)\citep{kenton2019bert, taylor1953cloze} and Translation Language Modeling (TLM)\citep{conneau2019cross} within a unified neural network architecture, which allows them to not only support multiple languages but also transfer knowledge across languages, thereby alleviating disparities in language processing.

\begin{figure}[pos=ht]
\centering
\includegraphics[width=0.8\textwidth,height=0.39\textwidth]{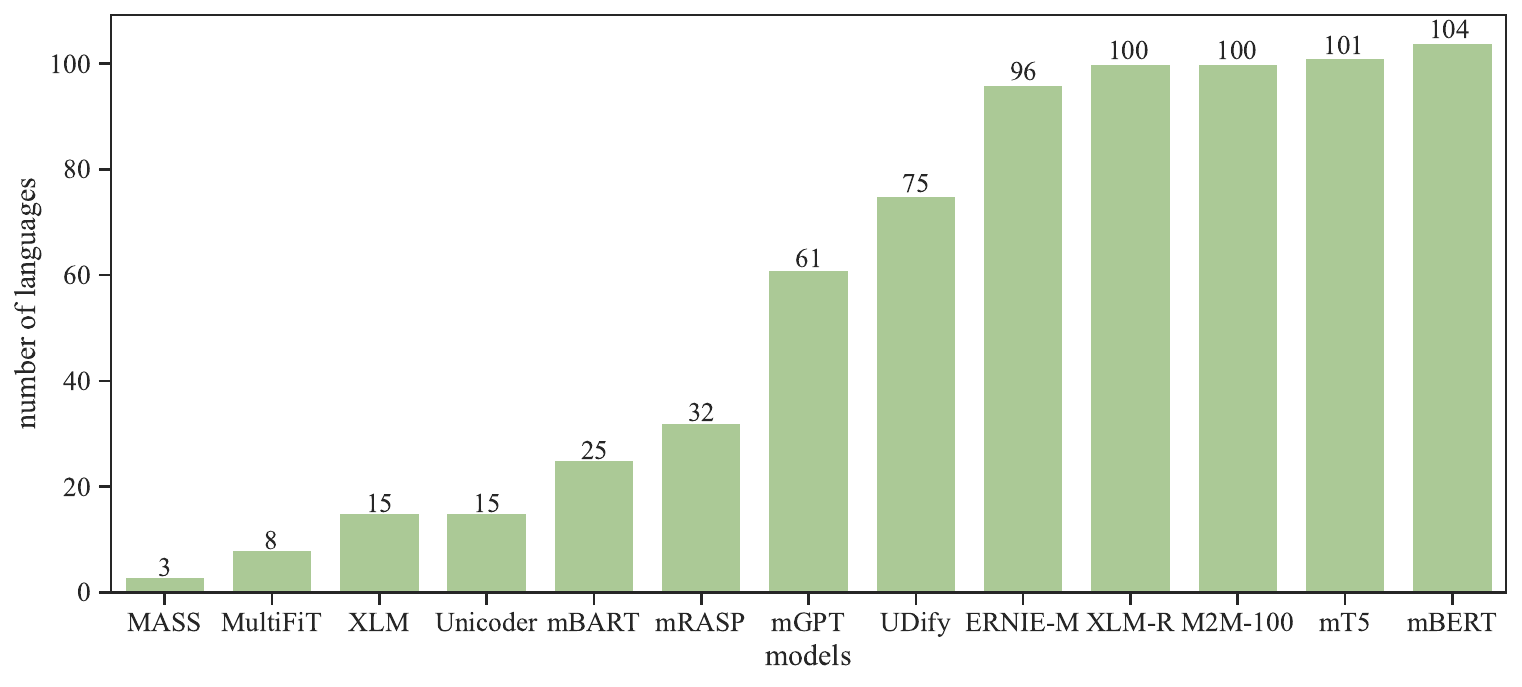} 
\caption{The number of languages supported by commonly used mPLMs.}
\label{fig.1}
\end{figure}

Currently, the number of languages supported by mPLMs remains relatively limited. Figure~\ref{fig.1} illustrates the number of languages supported by commonly used mPLMs\citep{kenton2019bert, conneau2020unsupervised, kale2020mt5, qinmass, eisenschlos2019multifit, huangunicoder, liu2020multilingual, shliazhko2024mgpt, kondratyuk201975, ouyang2021ernie, fan2021beyond}. It is evident that most of these models support only about one hundred languages, which is a very small proportion (roughly 1.5\%) compared to the nearly 7,000 languages\citep{campbell2008ethnologue} in the world. Furthermore, those models perform poorly on low-resource languages, particularly the languages with small corpus or significant grammatical and orthographic differences from high-resource languages (e.g. English)\citep{ebrahimi2021adapt}. Therefore, many works focus on the continued pre-training of pre-trained language models (PLMs) to better support low-resource languages\citep{wang2022expanding, ebrahimi2021adapt, dobler2023focus}. For example, Wang et al.\citep{wang2020extending} expand the vocabulary and continue to pre-train the multilingual BERT with the LORELEI corpus, enabling it to process previously unsupported languages, and demonstrate the effectiveness of their method in the named entity recognition (NER) task; Similarly, Ebrahimi et al.\citep{ebrahimi2021adapt} use the Bible corpus\citep{mccarthy2020johns} to extend models to support over 1,600 languages by expanding the vocabulary, continued pre-training, and incorporating language adapters\citep{rebuffi2017learning}, which improves the model's performance across many languages.

However, most prior work uses random initialization to represent the target language vocabulary. Typically, the parameters of vocabulary layer in pre-trained language models (PLMs) account for a significant portion of the model's total parameters. For example, in multilingual BERT (mBERT), the vocabulary layer accounts for approximately 52\% of the entire model's parameters\citep{abdaoui2020load}. Therefore, when continuing to pre-train mPLMs, the initial representation of the target language vocabulary directly impacts the convergence speed during pre-training and the final performance of the model \citep{dobler2023focus, minixhofer2022wechsel}. To address this issue, some previous works initialize the representation of target language vocabulary by aligning source and target word embeddings\citep{ruder2019survey} and calculating cross-lingual word similarities\citep{dobler2023focus, tran2019english}. However, these works typically focus on low-resource languages already supported by mPLMs\citep{dobler2023focus}, or on the extension of monolingual models\citep{vernikos2021subword}. In contrast, our work aims to process languages that the model does not support previously, where the models are relatively unfamiliar with the writing systems and grammar of these target languages. Additionally, our work focuses on multilingual models, rather than monolingual ones. Since most of the vocabulary in mPLMs is shared across languages, which encodes information involving multiple languages, it becomes more difficult to initialize the representation of target language vocabulary from the source language.

To address this issue, we need to expand the model's vocabulary to better support  the target language. To better initialize the representation of expanded vocabulary, we first screen out a vocabulary set that tends to represent the source language within mPLMs. Next, using the bilingual dictionary, we train cross-lingual (sub)word embeddings on large-scale monolingual corpora of the source and target languages, and compute the cross-lingual (sub)word similarity matrix. Subsequently, using the (sub)word representation of source languages in mPLMs and the cross-lingual similarity matrix, we calculate the initial representation of the expanded vocabulary for the target language through a weighted average. Finally, based on the representations of these expanded vocabulary, we continue to pre-train for the mPLMs on the target language corpus. The experimental results show that our proposed method outperforms other baseline methods in downstream tasks. In particular, for the NER task, the average performance of our method improves by 2.60\% compared to the baseline. Additionally, the method exhibits high robustness in selecting the training corpora, and the model's ability to process English does not significantly degrade after continued pre-training.

In this work, our contributions are as follows:
\begin{itemize}
\item [(1)] 
We propose a novel method for effectively initializing the embeddings of expanded vocabulary during the continued pre-training of multilingual pre-trained language models for low-resource languages. The expanded vocabulary provides better support and representation for these low-resource languages.
\item [(2)]
The mPLMs extended by our method demonstrate their effectiveness in POS tagging and NER tasks, outperforming the baseline by 0.54\% and 2.60\%, respectively.
\item [(3)]
We will release our extended models for the future research in the NLP community.
\end{itemize}

\section{Related work}
\subsection{Continued pre-training for pre-trained language models}
Due to limitations in data resources, time costs, and other factors, it is generally challenging to train a PLM from scratch for low-resource languages. As a result, many work attempt to perform continued pre-training on existing models to better support low-resource languages.

Regarding continued pre-training of monolingual language models, Tran\citep{tran2019english} adds an additional vocabulary layer for the target language, and concatenates it with the existing PLM. Then, to initialize the representation of the vocabulary layer, they use the corpora from both the source and target languages to learn cross-lingual representations. The model is then further trained using the corpora from both languages, with each language's respective vocabulary layer, resulting in a bilingual model that is capable of processing both the source and target languages. To alleviate the dependency on shared vocabulary during training cross-lingual word embeddings in Tran's work\citep{tran2019english}, Vernikos et al.\citep{vernikos2021subword} select word pairs with high alignment confidence as anchor words, which are then used to learn the word representations for the target languages. Additionally, some works extend to train monolingual language models to make them fully suitable for the target language. For instance, Vries et al.\citep{de2021good} continue to pre-train the GPT-2, making it suitable for both Dutch and Italian. Specifically, they fix the parameters of the model's encoding layers, and then train a new vocabulary layer using the target language corpus, which is concatenated with the existing encoding layers. To better initialize the target language vocabulary during the continued pre-training, Minixhofer et al.\citep{minixhofer2022wechsel} compute the similarity of cross-lingual word embeddings in both the source and target languages, thereby providing a better initialization for the target language vocabulary. They then continue to pre-train the model, making it suitable to the target language. Experimental results show that models trained in this way achieve better performance on tasks related to the target language.

Some other works focus on continued pre-training of mPLMs. Compared to monolingual language models, the continued pre-training of multilingual models can better leverage the existing multilingual knowledge within the model, enabling it to support the target language effectively\citep{wang2020extending, wu2020all}. For example, Wang et al.\citep{wang2019improving} first represent the target language vocabulary using joint mapping and composite mapping, and then continue to pre-train the mBERT model, which allows it to better process out-of-vocabulary (OOV) words in the target language. The trained model also achieves better performance on tasks such as POS tagging and NER. Additionally, Wang et al.\citep{wang2022expanding} use bilingual dictionaries to replace words from the source language with their target language counterparts, constructing pseudo-target sentences. They then continue to pre-train the mBERT model, enabling it to process new languages that are not previously supported. Ebrahimi et al.\citep{ebrahimi2021adapt} employ techniques such as continued pre-training, vocabulary expansion, and adapters with Bible corpora\citep{mccarthy2020johns} to extend the XLM-R for low-resource languages. Dobler et al.\citep{dobler2023focus} first train word embeddings using the target language corpus. They then select out the shared vocabulary between the target language and the vocabulary in mBERT model. Using these shared vocabularies, they initialize the word embeddings for the target language. Finally, they continue to pre-train the mBERT model to better support new target language.

Our work is inspired by Wang et al.'s\citep{wang2020extending}. They train a target language vocabulary using LORELEI corpora\citep{strassel2016lorelei}, and integrate it into the existing vocabulary of the mBERT model. Next, they continue to pre-train the model with the LORELEI corpora to enable it to support the target language. However, in their work, the expanded vocabulary for the target language is initialized randomly, which limits the final training performance and increases training time. Therefore, our work calculates the cross-lingual similarity based on bilingual (sub)word embeddings from the source and target languages, providing a better initialization for the target language vocabulary.

\subsection{Multilingual representation learning}
In multilingual representation learning, a common approach is to train static word embeddings for each language separately, and map the word embeddings of the target languages into the embedding space of the source language, using bilingual dictionaries between the source language (usually English) and the other languages\citep{ammarmassively}. However, this method aligns the word representations of each target language with the source language independently, leading to suboptimal alignment performance among those target languages. Therefore, to improve multilingual word representations, the following work\citep{jawanpuria2019learning} employs the generalized procrustes analysis\citep{kementchedjhievao2018generalizing}, which jointly trains corpora from multiple languages, and aligns them into a new high-dimensional embedding space.

In context-based multilingual representation learning, a common approach is to apply self-supervised learning on multiple languages' corpus using the same model architecture, such as mBERT\citep{kenton2019bert}, XLM-R\citep{conneau2020unsupervised}, and mT5\citep{kale2020mt5}. Although many mPLMs do not explicitly incorporate cross-lingual alignment signals during training, words from different languages are implicitly represented in the same embedding space through a shared vocabulary and a unified model architecture.

\section{Method}
\begin{figure}[pos=ht]
\centering
\includegraphics[width=0.9\textwidth,height=0.55\textwidth]{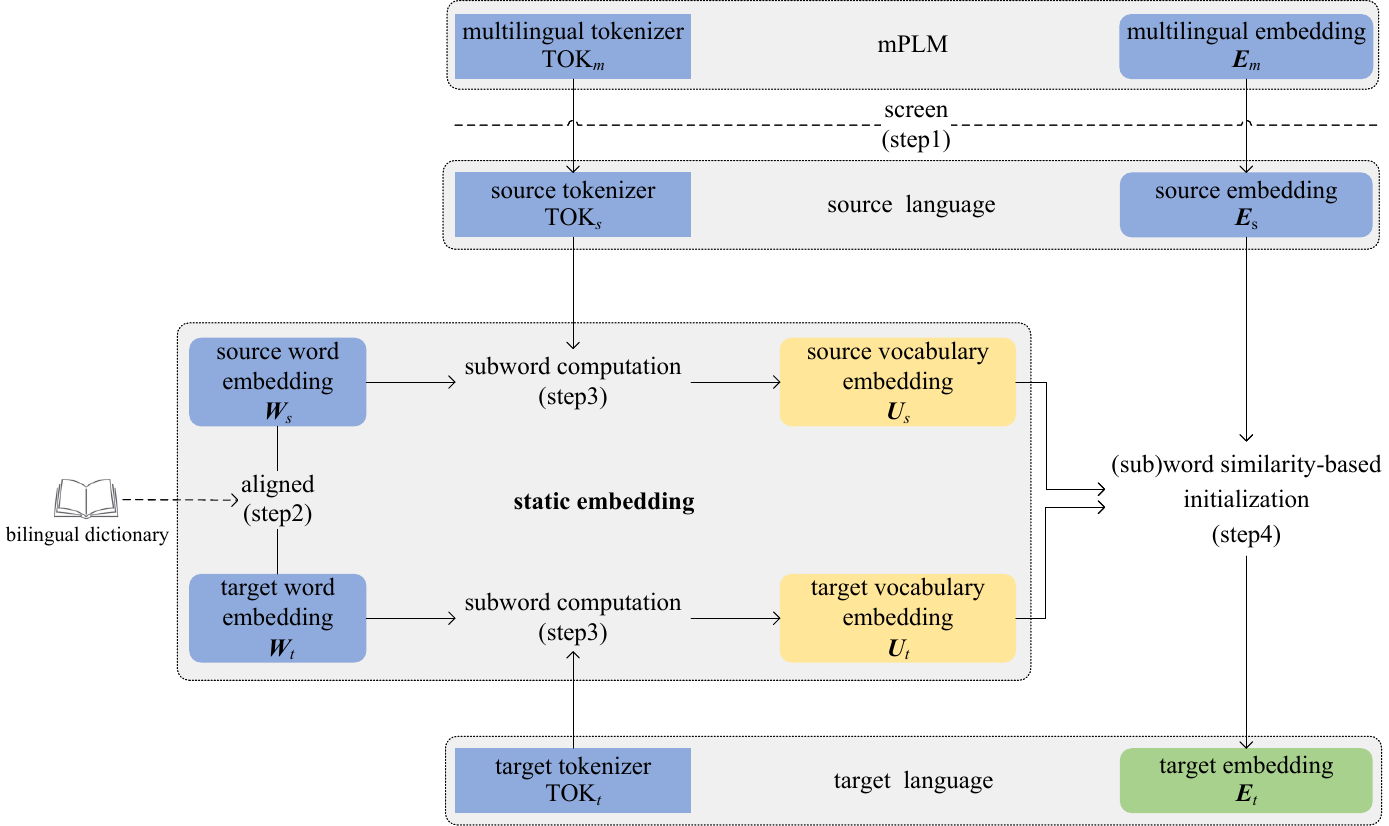} 
\caption{The steps for initializing the representations of the expanded vocabulary for target languages.}
\label{fig.2}
\end{figure}

Our aim is to improve the initial representation of the expanded vocabulary for the target language during the continued pre-training of mPLMs. As shown in Figure~\ref{fig.2}, we first screen out the word set ${V}_{s}$ from the multilingual model's vocabulary which is biased toward representing the source language, and obtain their embeddings $\boldsymbol{E}_s$ (Step 1). Then, using the bilingual dictionary, we align the monolingual word embeddings for both the source language $\boldsymbol{W}_s$ and target language $\boldsymbol{W}_t$ (Step 2). Next, to represent the target language subwords in the expanded vocabulary, we calculate the subword embeddings  according to the monolingual word embeddings, following fastText's approach\citep{bojanowski2017enriching} for training out-of-vocabulary (OOV) words (Step 3). Finally, by calculating the similarity matrix $\mathbf{S}$ between the source and target language vocabulary embeddings, we obtain the target language's embeddings $\boldsymbol{E}_t$ through weighted averaging of the corresponding source language embeddings $\boldsymbol{E}_s$ (Step 4).

\subsection{Screening source language vocabulary }
In mPLMs, most (sub)words in the vocabulary are shared and encode information across multiple languages. As a result, aligning the entire vocabulary of the model with the expanded vocabulary for the target language would incur substantial computational costs. To mitigate this, we assume that vocabularies from different languages in multilingual models can be approximately represented within a shared embedding space. However, due to discrepancies in the size of training corpora across languages, the model tends to represent high-resource languages (e.g., English) more effectively, allocating a substantial portion of the vocabulary to support these languages. Considering this, we propose to select a high-resource language as the source language and screen out a subset of the mPLM's vocabulary, which is better suitable for representing the high-resource language, and obtain their static embeddings.

During screening, we assume that there exists a monolingual PLM specifically trained for the high-resource language. This monolingual model is trained on large-scale corpora of this language, and its vocabulary ${V}_{s1}$ being well set and represented. Hence, we can screen out a subset from the multilingual model's vocabulary ${V}_{m}$ that overlaps the monolingual model's vocabulary, which is used to represent the source language, that is ${V}_{s} = {V}_{s1}\cap{V}_{m}$.

\subsection{Training cross-lingual word embeddings}
Using large-scale monolingual corpora, we train word embeddings for both the source and target languages, denoted as $\boldsymbol{W}_s$ and $\boldsymbol{W}_t$, respectively. Then, utilizing the bilingual dictionary, we adopt the orthogonal method\citep{schonemann1966generalized} to align the word embeddings of the source and target languages\footnote{In cases where a bilingual dictionary is unavailable for the language pair, we can alternatively train cross-lingual word embeddings in an unsupervised manner}\citep{lample2018word}. Notably, when aligning word embeddings across different languages, we relax the assumption of isomorphism between the embedding spaces of the source and target languages\citep{vulic2020all}.

\subsection{Calculating subword embeddings}
In the field of NLP, a subword refers to a linguistic unit that is smaller than a word but larger than a character. In this work, considering that some subwords in the expanded vocabulary for the target language need to be represented, we propose to compute the corresponding subword embeddings based on the word embeddings.

Specifically, given the tokenizers for the source language and target language, denoted as TOK$_{s}$ and TOK$_{t}$, respectively, we obtain the vocabularies for both languages, denoted as ${V}_{s}$ and ${V}_{t}$. To represent the subwords in these vocabularies, we refer to the fastText's approach\citep{bojanowski2017enriching} for representing out-of-vocabulary (OOV) words. The subword representations can be computed as follows:
$$\boldsymbol{u}_z=\displaystyle\sum_{g\in \zeta (z)}\boldsymbol{w}_g$$
\noindent{Where, $g\in \zeta (z)$ denotes the set of all n-grams in the subword $z$, $\boldsymbol{w}_g$ is the word embedding of a specific n-gram fragment in subword $z$, and $\boldsymbol{u}_z$ is the embedding of the subword $z$. If none of the n-gram combinations for a subword exist in the word sets, the embedding representation for that subword is set to a zero embedding. }

Through this method, these subword embeddings can be represented in the same space as the corresponding word embeddings. Finally, we will combine the word and subword representations to obtain the vocabulary representations for both the source language and the target language, denoted as $\boldsymbol{U}_s$ and $\boldsymbol{U}_t$, respectively.

\subsection{Initializing the expanded vocabulary's representation for target languages based on (sub)word similarity}
We first compute the similarity matrix $\mathbf{S}$ between the source language vocabulary embeddings $\boldsymbol{U}_s$ and the target language vocabulary embeddings $\boldsymbol{U}_t$, as shown in the following equation:
$$\mathbf{S}=\frac{\boldsymbol{U}_s\boldsymbol{U}_t}{\left \|\boldsymbol{U}_s\right \|\left \|\boldsymbol{U}_s\right \|}$$

Then, for each target language (sub)word, based on the similarity matrix $\mathbf{S}$, we select the $k$ most similar (sub)words from the source language vocabulary. Using these  (sub)words' representations in source language vocabulary, a weighted average is computed on their similarity to obtain the target language (sub)word representation:
$${e}_{y}^{t}=\frac{\textstyle\sum_{x\in \varpi (y)}exp({S}_{x,y}/\tau)\cdot {e}_{x}^{s}}{\textstyle\sum_{x'\in \varpi (y)}exp({S}_{x,y}/\tau)}$$

Where, ${e}_{y}^{t}$ denotes the target embedding for (sub)word $y$ in the expanded vocabulary, while ${e}_{x}^{s}$  represents the source embedding for (sub)word $x$ in the mPLM's vocabulary. $\varpi (y)$ denotes the set of the $k$ source (sub)words that are most similar to the target  (sub)word $y$, as determined by the similarity matrix $\mathbf{S}$. ${S}_{x,y}$ represents the similarity score between the source language (sub)word $x$ and the target language (sub)word $y$ through $\mathbf{S}$, and $\tau$ is a smoothing factor. In addition, if ${e}_{y}^{t}$ is a zero vector after computation, it will be initialized by sampling from a normal distribution estimated from all source embeddings, that is $N(E[\boldsymbol{E}_s], Var[\boldsymbol{E}_s])$.

Through the above steps, the initial representation for the target language vocabulary, $\boldsymbol{E}_t$, will be approximately in the same embedding space as the vocabulary representations in mPLM model. Additionally, to preserve the original multilingual characteristics of the model, we only initialize the non-overlapping vocabulary in the expanded vocabulary for the target language with the original vocabulary in the mPLM model.

\section{Experiment}
\subsection{Dataset}
We select English as the source language. This is because English, as a global lingua franca, has a wealth of available corpora and processing tools. Besides, the existing mPLMs represent and process English effectively, resulting in a significant proportion of the model's vocabulary being biased to represent English. Considering that the source word embeddings should be of high quality, we use the pre-trained English word embedding provided by fastText\footnote{https://fasttext.cc/docs/en/crawl-vectors.html} as the source language word representation.

Considering the scale of the corpus for the target languages and the availability of evaluation datasets, we select seven low-resource languages as the target languages, shown in Table~\ref{table.1}. Additionally, we adopt MADLAD-400 dataset\footnote{https://huggingface.co/datasets/allenai/MADLAD-400} to train monolingual word embeddings, generate expanded vocabulary, and perform continued pre-training on mPLMs.

\begin{table}[pos=h]
\caption{The introduction of target language corpus.}
\label{table.1}
\begin{tabular}{cccc}
\hline
\textbf{Target languages} & \textbf{Language family-Language group} & \textbf{Sentence counts} & \textbf{Token counts} \\
\hline
Amharic (am)& Afro-Asiatic—Semitic& 3.25M& 98.37M\\
Maltese (mt)& Afro-Asiatic—Semitic& 6.44M& 177.69M\\
Manx (gv)& Indo-European—Celtic& 38.37K& 1.03M\\
North\_Sami (se)& Uralic—Finno-Ugric& 508.50K& 10.51M\\
Scottish (gd)& Indo-European—Germanic& 2.63M& 87.01M\\
Uyghur (ug)& Altaic—Turkic& 2.04M& 64.60M\\
Sinhala (si)& Indo-European—Indic& 10.78M&303.72M\\
\hline
\end{tabular}
\end{table}

\subsection{Tools}
\noindent{\textbf{Base Model}  Considering the generality and representativeness of the base model, we select the multilingual BERT (mBERT)\citep{kenton2019bert} as the base model for continued pre-training. This model is primarily trained on Wikipedia corpora and uses the WordPiece algorithm\footnote{https://github.com/google-research/bert/} to generate its vocabulary. Masked Language Modeling (MLM)\citep{kenton2019bert, taylor1953cloze} and Next Sentence Prediction (NSP)\citep{kenton2019bert} are employed as pre-training tasks. mBERT supports 104 languages in total. Additionally, the model has a vocabulary size of 119,547 and an embedding dimension of 768.}

\noindent{\textbf{Training tools for word embeddings}  We use fastText\citep{bojanowski2017enriching} to train the monolingual word embeddings for each target language. FastText employs a single-layer neural network architecture based on the continuous bag-of-words (CBOW) model. During training, words in the text are represented as feature embeddings, which are then input into the neural network for linear computation. Finally, the category label probabilities of the input text are obtained through a softmax layer.}

\noindent{\textbf{Bilingual dictionary}  When aligning the static word embeddings of the source and target languages, we use the bilingual dictionary developed by Minixhofer et al.\citep{minixhofer2022wechsel}. The dictionary is constructed by crawling entries from Wikipedia, and currently supports over 3,200 language pairs. The scale of bilingual word pairs used in this work is shown in Table~\ref{table.2}.}

\noindent{\textbf{Tokenizer}  To better adapt to the existing mBERT model, we adopt the WordPiece algorithm to generate the target language vocabulary, which then serves as the tokenizer.}

\begin{table}[pos=h]
\caption{The scale of bilingual dictionary for target languages.}
\label{table.2}
\begin{tabular}{cccccccc}
\hline 
\textbf{Target languages}& \begin{tabular}[c]{@{}c@{}}Amharic\\ (am)\end{tabular} & \begin{tabular}[c]{@{}c@{}}Maltese\\ (mt)\end{tabular} & \begin{tabular}[c]{@{}c@{}}Manx\\ (gv)\end{tabular} & \begin{tabular}[c]{@{}c@{}}North\_Sami\\ (se)\end{tabular} & \begin{tabular}[c]{@{}c@{}}Scottish\\ (gd)\end{tabular} & \begin{tabular}[c]{@{}c@{}}Uyghur\\ (ug)\end{tabular} & \begin{tabular}[c]{@{}c@{}}Sinhala\\ (si)\end{tabular} \\
\hline 
\textbf{Sentence pair counts} & 0.8k& 4.5k& 5.7k& 3.3k& 8.2k& 1.1k& 0.2k\\           \hline                                       
\end{tabular}
\end{table}

\subsection{Baselines}
\noindent{\textbf{Multilingual BERT model}  We directly use the existing multilingual BERT model to perform zero-shot cross-lingual transfer for these target languages. We refer to this baseline as \textit{mBERT}.}

\noindent{\textbf{Random initialization for the expanded vocabulary during continued pre-training}  After expanding the base model's vocabulary using target language corpus, we randomly initialize the representations of these vocabulary, and then continue to pre-train the base model with the target language corpus. We refer to this baseline as \textit{Random initialization}.}

\subsection{Evaluation tasks}
Since the target languages in this experiment are all low-resource languages, various evaluation datasets for these languages are lack. We select POS tagging and NER tasks for evaluation, as they support a wide range of languages. Specifically, we use the Universal Dependency (UD) v2.13\footnote{https://universaldependencies.org/introduction.html} dataset for the POS tagging task. This dataset adopt a unified annotation scheme, including 17 tags, such as nouns, verbs, and adjectives, and support the seven target languages in our experiment. For the NER task, we select wikiann\footnote{https://huggingface.co/datasets/unimelb-nlp/wikiann}, a dataset collected from Wikipedia corpora, which includes three tags: person, location, and organization. This dataset support five of the target languages. Table~\ref{table.3} shows the annotation scale for each language in the two tasks.

Furthermore, for comparison with other baselines, we adopt the evaluation script from XTREME\citep{siddhant2020xtreme}, which fine-tunes the BERT model directly, allowing it to output the predicted label for each word. In our experiment, the evaluation metric for both tasks is the macro-average F1 score.

\begin{table}[pos=h]
\caption{The statistics of evaluation tasks for the target languages.}
\label{table.3}
\begin{tabular}{cccclccc}
\hline
\multirow{2}{*}{\textbf{Target language}} & \multicolumn{3}{c}
{\textbf{POS tagging}} & \multicolumn{1}{c}{} & \multicolumn{3}{c}{\textbf{NER}} \\
\cline{2-8}
& train& validation& test& \multicolumn{1}{c}{}& train& validation& test\\
\hline
Amharic (am)& /& /& 1,073&& 100& 100& 100\\
Maltese (mt)& 1,123& 433& 518 && 100& 100& 100\\
Manx (gv)& 1,172& /& 1,164&& /& /& /\\
North\_Sami (se)& 2,257& /& 865&& /& /& /\\
Scottish (gd)& 3,541& 655& 545&& 100& 100& 100\\
Uyghur (ug)& 1,656& 900& 900&& 100& 100& 100\\
Sinhala (si)& /& /& 99&& 100& 100& 100\\
\hline
\end{tabular}
\end{table}

\subsection{Parameter setup}
When using fastText to train the monolingual word embeddings $\boldsymbol{W}_s$ and $\boldsymbol{W}_t$, we employ the CBOW method, and set the word embedding dimension to 300, training epochs to 20, and the negative sampling to 10, with all other parameters to the default values. Additionally, for the parameters mentioned in section 3.4, we set the number of nearest neighboring words $k$ to 10; the smoothing factor $\tau$ to 0.1. During the continued pre-training of mBERT, the specific training parameters are listed in Table~\ref{table.4}. Finally, when performing the evaluation tasks using the continued pre-training models, we adjust the training epochs to 20, with all other parameters following the default values.

\begin{table}[pos=h]
\caption{The parameters for continued pre-training of mPLMs.}
\label{table.4}
\begin{tabular}{cc}
\hline
\textbf{parameters} & \textbf{values} \\
\hline
vocabulary size& 30k\\
step& $\sim$500k\\
learning rate& 2e-5\\
training task& MLM\\
tokenizer& WordPiece\\
optimizer& adamw\\
batch size& 32\\
word length& 128\\ 
\hline
\end{tabular}
\end{table}

\section{Results and analysis}
\subsection{Main experimental results}
The main experimental results are shown in Table~\ref{table.5}. Our proposed method outperforms other baselines in most of target languages. In the POS tagging and NER tasks, compared to \textit{Random initialization} baseline, the average performance improves by 0.54\% and 2.60\%, respectively. Notably, in the POS tagging task, our proposed method does not performs well with the \textit{Random initialization} baseline for Amharic (am) and Sinhala (si). This may be because both languages have fewer than 1,000 bilingual word pairs, impacting the cross-lingual alignment. As a result, the expanded vocabulary for the target language could not be effectively initialized using our method.

\begin{table}[pos=h]
\caption{The performance of each method on evaluation tasks.}
\label{table.5}
\begin{tabular}{cccccccccc}
\hline
\textbf{Task}& \textbf{method}& \textbf{am}& \textbf{gd}& \textbf{gv}& \textbf{mt}& \textbf{se}& \textbf{si} &\textbf{ug}& \textbf{ave}\\
\hline
\multirow{3}{*}{\begin{tabular}[c]{@{}c@{}}POS\\ tagging\end{tabular}} & mBERT& 0.02& 40.37& 31.36& 23.84& 32.22& 5.68& 14.07& 21.08\\
& Random initialization & \textbf{40.26} & 57.34 & 63.11& 72.90& 63.20& \textbf{29.63} &48.87 &53.62\\
& ours& 37.62& \textbf{57.62}& \textbf{68.67}& \textbf{73.23}& \textbf{63.97}& 28.79& \textbf{49.25}& \textbf{54.16} \\
\hline
\multirow{3}{*}{NER}& mBERT& 0& 48.39& /& 50.95& /& 0& 13.48& 22.56\\
& Random initialization& 41.04& 49.08& /& 56.20& /& 25.00& 33.72& 41.01\\
& ours& \textbf{42.15}& \textbf{50.92}& /& \textbf{62.88}& /& \textbf{25.44}& \textbf{36.68} &\textbf{43.61}\\
\hline
\end{tabular}
\end{table}

Additionally, we find that even without using our proposed method, randomly expanding the models' vocabulary and continuing to pre-train the multilingual model with target language corpora can significantly improve the performance for the target language. As shown in Table~\ref{table.5}, the \textit{mBERT} baseline performs the worst on both tasks across all the languages. In particular, for Amharic (am), Sinhala (si), and Uyghur (ug), which differ significantly from the source language in terms of writing systems and language typologies, the performance on the two tasks is sometimes as low as 0. In contrast, the \textit{Random initialization} baseline shows significantly better performance than \textit{mBERT} baseline with average improvements of 32.54\% and 18.45\% in the POS tagging and NER tasks, respectively. Notably, for Amharic (am), the performance in the two tasks improves by 40.24\% and 41.04\%, respectively, which reflects the importance of vocabulary expansion and continued pre-training. Table~\ref{table.6} lists the number of vocabulary expanded for each target language, all of which exceed 21,000 words, with some languages reaching up to 25,000 words. This indicates that the original vocabulary in mBERT does not represent these target languages well. Therefore, after expanding the vocabulary for the target language, the mBERT model can better segment and represent the target language texts. Then, by continuing to pre-train the models with the target language corpus, the models can become more familiar with the grammar of the target language.

\begin{table}[pos=h]
\caption{The statistics of expanded vocabulary for the target languages.}
\label{table.6}
\begin{tabular}{cccccccc}
\hline
\textbf{Target language}                                            & \begin{tabular}[c]{@{}c@{}}Amharic\\ (am)\end{tabular} & \begin{tabular}[c]{@{}c@{}}Maltese\\ (mt)\end{tabular} & \begin{tabular}[c]{@{}c@{}}Manx\\ (gv)\end{tabular} & \begin{tabular}[c]{@{}c@{}}North\_Sami\\ (se)\end{tabular} & \begin{tabular}[c]{@{}c@{}}Scottish\\ (gd)\end{tabular} & \begin{tabular}[c]{@{}c@{}}Uyghur\\ (ug)\end{tabular} & \begin{tabular}[c]{@{}c@{}}Sinhala\\ (si)\end{tabular} \\
\hline
\textbf{\begin{tabular}[c]{@{}c@{}}Number of \\ expanded vocabulary\end{tabular}} & 25,376& 21,772& 22,068& 25,566& 21,210& 24,488& 24,213 \\         
\hline
\end{tabular}
\end{table}

For specific tasks, our proposed method demonstrates a significant improvement in the NER task. Compared to the \textit{Random initialization} baseline, our method yields a consistent improvement, achieving an average gain of 2.60\%, with the largest improvement reaching 6.68\%. This may be because, unlike the POS tagging task, the NER task requires to predict fewer labels, making it easier to achieve a significantly improved performance.

Finally, we hypothesize that the relationship between the source language and the target language may influence the models' performance to some extent. For example, for Scottish (gd), which belongs to the same language family as the source language, English, our proposed method shows minor improvements. In contrast, for languages that are more distantly related to the source language, such as Maltese (mt) and Uyghur  (ug), the improvement is significant. This may be because, when the target language shares more similarities with the source language in writing systems and linguistic features, the base model can still process it effectively, even if it has not seen the target language during pre-training. In contrast, if the target language is less closely related to the source language, which means the model has less exposure to and learned that language during pre-training. Therefore, using our proposed method, the model's performance can achieve a larger improvement for such languages.

\subsection{Analysis}
\subsubsection{The impact of vocabulary size of source language}
\begin{figure}[pos=ht]
\centering
\includegraphics[width=0.85\textwidth,height=0.45\textwidth]{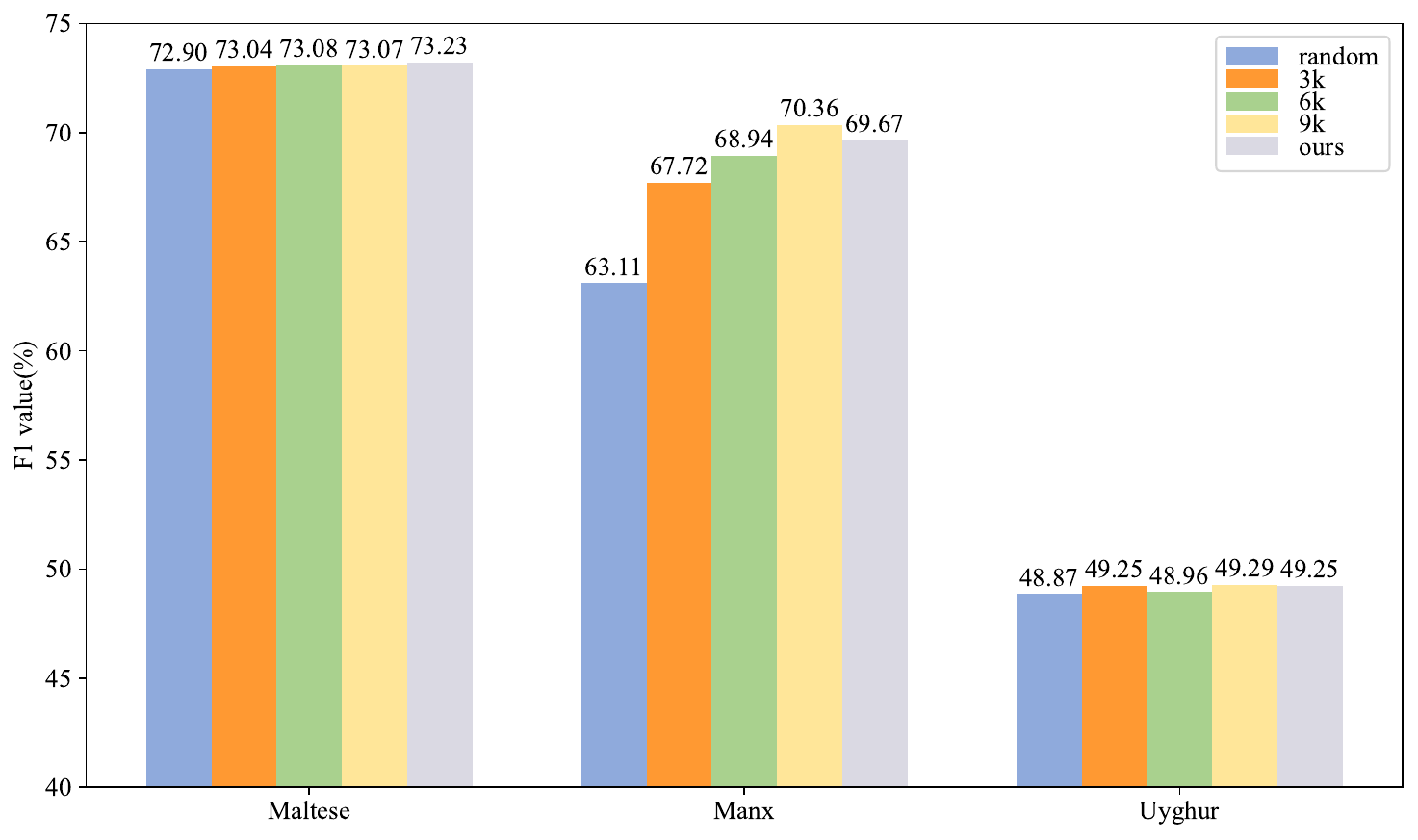} 
\caption{The impact of source language vocabulary size on the POS tagging task.}
\label{fig.3}
\end{figure}

\begin{figure}[pos=ht]
\centering
\includegraphics[width=0.7\textwidth,height=0.38\textwidth]{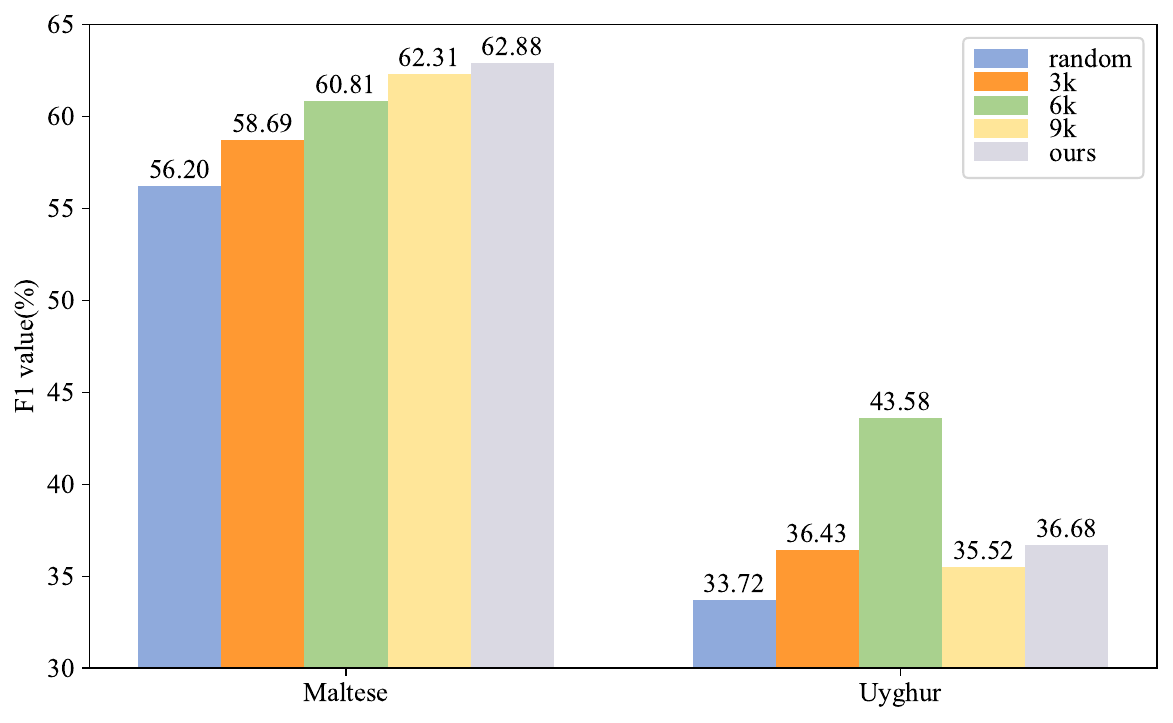} 
\caption{The impact of source language vocabulary size on the NER task.}
\label{fig.4}
\end{figure}

After screening out the vocabulary in section 3.1, we find that over 13k (sub)words biased toward representing the source language are selected. According to these vocabulary, we investigate whether the vocabulary size screened out affects the performance of the trained models. Specifically, we randomly select 3k, 6k, and 9k (sub)words from the source language vocabulary set and continue to pre-train the models with our method. The performance of the extended models on downstream tasks is shown in Figure ~\ref{fig.3} and ~\ref{fig.4}. According to the two figures, we find that our method still outperforms the \textit{Random initialization} baseline as long as the source language vocabulary set reaches a certain size. Additionally, we observe that a larger source language vocabulary set does not necessarily lead to better performance. This may be because the quality of the vocabulary set also affects the models' performance, which inspires us to design more precise measure to screen out a high-quality source language vocabulary set in future work.

\subsubsection{The impact of training corpus}
We also investigate whether the size of the target language corpus affects the models' performance. Combing the results in Table~\ref{table.1} and Table~\ref{table.5}, we find no significant correlation between the two. This may be because, once a sufficient training corpus is provided, the trained models can become familiar with the target language's vocabulary representations and grammatical rules through vocabulary expansion and continued pre-training.

Additionally, we explore the impact of the source of the target language corpus on models' performance. Specifically, we replace the training corpus from MADLAD-400 to Wikipedia (wiki) data\footnote{https://dumps.wikimedia.org/}, and continue to pre-train the models for three target languages: Maltese (mt), Manx (gv), and Uyghur (ug). In the Wikipedia dataset, the corpus sizes of these three languages are 1.08M, 22.68K, and 91.41K sentences, respectively. After continued pre-training with the Wikipedia corpus, the models' performance on downstream tasks is shown in Table~\ref{table.7}. According to Table~\ref{table.7}, we find that whether the training corpus is MADLAD-400 or Wikipedia, our proposed method consistently outperforms the \textit{Random initialization} baseline. This suggests that, regardless of the training corpus source, our method is effective. Moreover, we also observe that the source of the training corpus does impact on the models' performance on specific downstream tasks. The models trained with MADLAD data consistently outperform those trained with Wikipedia data on the POS tagging task, while the opposite performance is observed for the NER task. This indicates that the choice of training corpus should consider the specific downstream tasks.

\begin{table}[pos=h]
\caption{The expanded models' performance trained on corpora with various sources.}
\label{table.7}
\begin{tabular}{cccccccc}
\hline
\multirow{2}{*}{\textbf{Task}} & \multirow{2}{*}{\textbf{Method}}& \multicolumn{2}{c}{Maltese (mt)}& \multicolumn{2}{c}{Manx (gv)}& \multicolumn{2}{c}
{\begin{tabular}[c]{@{}c@{}}Uyghur (ug)\end{tabular}} \\
\cline{3-8}
& & MADLAD& wiki& MADLAD& wiki& MADLAD& wiki\\
\hline
\multirow{2}{*}{POS tagging}& \begin{tabular}[c]{@{}c@{}}Random\\ initialization\end{tabular} & 72.90& 70.53& 48.87& 46.78& 63.11& 64.82\\
& ours& \textbf{73.23}& \textbf{70.77}& \textbf{49.25}& \textbf{48.04}& \textbf{68.67}& \textbf{65.20}\\
\hline
\multirow{2}{*}{NER}& \begin{tabular}[c]{@{}c@{}}Random\\ initialization\end{tabular} & 56.20& 64.18& 33.72& 39.69& /& /\\
& ours& \textbf{62.88}& \textbf{65.03}& \textbf{36.68}& \textbf{43.18}& /& /\\                         
\hline
\end{tabular}
\end{table}

\subsubsection{The impact on source language:English}
\begin{figure}[pos=ht]
\centering
\includegraphics[width=0.55\textwidth,height=0.33\textwidth]{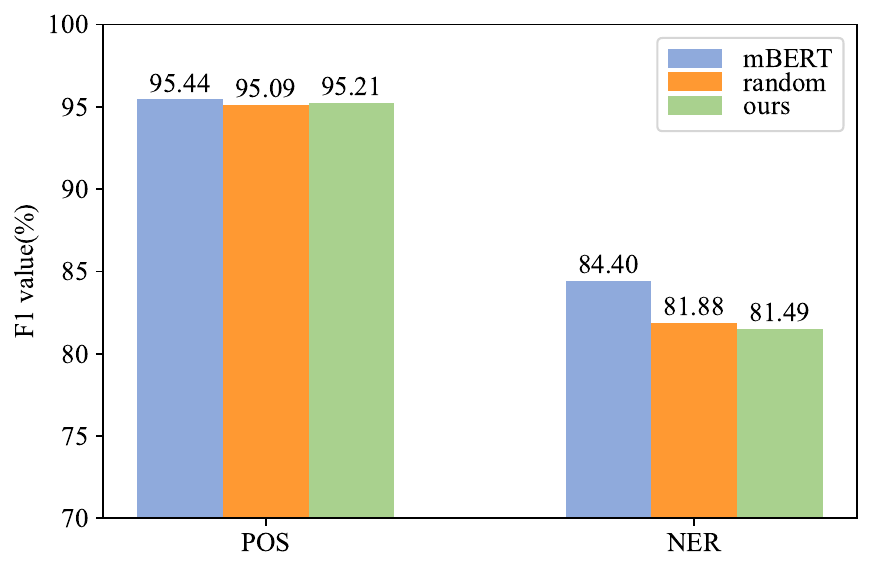} 
\caption{The performance of each method for English on the evaluation tasks.}
\label{fig.5}
\end{figure}

We also investigate, after continued pre-training, whether the extended models' performance changes when processing the English tasks. The experimental results are shown in Figure~\ref{fig.5}. For both the \textit{Random initialization} baseline and our proposed method, we compute the arithmetic average of the performance of the models trained for all target languages on English tasks. As shown in Figure~\ref{fig.5}, after continued pre-training, the models' performance on the English POS tagging task shows almost no decline, while a slight drop in performance on the English NER task, but not particularly significant. This suggests that, even after continuing to pre-train the mBERT model with target language corpora, the trained models still retain a strong grasp of English knowledge.

\subsubsection{Limitation and challenge}
In this work, we use bilingual dictionaries to expand the vocabulary for the target language. However, the performance of the extended models will be affected if the bilingual dictionary is small. Hence, it is necessary to construct and use larger,  higher-quality bilingual dictionaries. Additionally, since this work focuses on extremely low-resource languages, which often lack diverse evaluation datasets, such as those for reading comprehension, semantic parsing, even document classification tasks, which further highlights the need to develop evaluation resources tailored to extremely low-resource languages in the future. Finally, the quality of the source language vocabulary set still requires further improvement. Our future work will focus on designing more precise methods to select a more representative source language vocabulary set.

\section{Conclusion}
In this work, to enable existing mPLMs to better process low-resource languages, we first expand the models' vocabulary to include these languages. Then, using the source language's word representations and word alignment provided by bilingual dictionaries, we better initialize the representation of the expanded vocabulary for the target language. The experimental results show that our proposed method outperforms the \textit{Random initialization} baseline. In the future, we will design new methods for extending mPLMs to simultaneously support the processing of multiple extremely low-resource languages.

\vspace{1em}
\noindent{\textbf{Data Availability Statement}: We release the code and data at: https://github.com/JianyuZheng5/Vocabulary-expanding-of-multilingual-language-models.}

\vspace{1em}
\noindent{\textbf{Acknowledgement}: I thank Ying Liu at Tsinghua University for her helpful comments. I also thank the Center of High performance computing, Tsinghua University for providing computing resources.}

\vspace{1em}
\noindent{\textbf{Funding Statement}: This work was supported by Fundamental Research Funds for the Central Universities of Ministry of Education of China(Grant No. ZYGX2025WXJ007), the Postdoctoral Fellowship Program of CPSF (Grant No. GZC20252558), and the China Postdoctoral Science Foundation (Grant No. 2025M783842).}

\printcredits

\bibliographystyle{cas-model2-names}

\bibliography{cas-refs}

@article{doddapaneni2025primer,
  title={A primer on pretrained multilingual language models},
  author={Doddapaneni, Sumanth and Ramesh, Gowtham and Khapra, Mitesh and Kunchukuttan, Anoop and Kumar, Pratyush},
  journal={ACM Computing Surveys},
  volume={57},
  number={9},
  pages={1--39},
  year={2025},
  publisher={ACM New York, NY}
}

@inproceedings{kenton2019bert,
  title={Bert: Pre-training of deep bidirectional transformers for language understanding},
  author={Kenton, Jacob Devlin Ming-Wei Chang and Toutanova, Lee Kristina and others},
  booktitle={Proceedings of naacL-HLT},
  volume={1},
  number={2},
  year={2019},
  organization={Minneapolis, Minnesota}
}

@article{taylor1953cloze,
  title={“Cloze procedure”: A new tool for measuring readability},
  author={Taylor, Wilson L},
  journal={Journalism quarterly},
  volume={30},
  number={4},
  pages={415--433},
  year={1953},
  publisher={SAGE Publications Sage CA: Los Angeles, CA}
}

@article{conneau2019cross,
  title={Cross-lingual language model pretraining},
  author={Conneau, Alexis and Lample, Guillaume},
  journal={Advances in neural information processing systems},
  volume={32},
  year={2019}
}

@inproceedings{conneau2020unsupervised,
  title={Unsupervised cross-lingual representation learning at scale},
  author={Conneau, Alexis and Khandelwal, Kartikay and Goyal, Naman and Chaudhary, Vishrav and Wenzek, Guillaume and Guzm{\'a}n, Francisco and Grave, Edouard and Ott, Myle and Zettlemoyer, Luke and Stoyanov, Veselin},
  booktitle={Proceedings of the 58th annual meeting of the association for computational linguistics},
  pages={8440--8451},
  year={2020}
}

@article{kale2020mt5,
  title={mT5: a massively multilingual pre-trained text-to-text transformer},
  author={Kale, M and Xue, L and Constant, N and Roberts, A and Al-Rfou, R and Siddhant, A and Barua, A},
  journal={arXiv preprint arXiv:2010.11934},
  year={2020}
}

@article{qinmass,
  title={MASS: Masked Sequence to Sequence Pre-training for Language Generation},
  author={Qin, Tao}
}

@inproceedings{eisenschlos2019multifit,
  title={MultiFiT: Efficient multi-lingual language model fine-tuning},
  author={Eisenschlos, Julian and Ruder, Sebastian and Czapla, Piotr and Kadras, Marcin and Gugger, Sylvain and Howard, Jeremy},
  booktitle={Proceedings of the 2019 conference on empirical methods in natural language processing and the 9th international joint conference on natural language processing (EMNLP-IJCNLP)},
  pages={5702--5707},
  year={2019}
}

@article{huangunicoder,
  title={Unicoder: A Universal Language Encoder by Pre-training with Multiple Cross-lingual Tasks},
  author={Huang, Haoyang and Liang, Yaobo and Duan, Nan and Gong, Ming and Shou, Linjun and Jiang, Daxin and Zhou, Ming}
}

@article{liu2020multilingual,
  title={Multilingual denoising pre-training for neural machine translation},
  author={Liu, Yinhan and Gu, Jiatao and Goyal, Naman and Li, Xian and Edunov, Sergey and Ghazvininejad, Marjan and Lewis, Mike and Zettlemoyer, Luke},
  journal={Transactions of the Association for Computational Linguistics},
  volume={8},
  pages={726--742},
  year={2020},
  publisher={MIT Press One Rogers Street, Cambridge, MA 02142-1209, USA journals-info~…}
}

@article{shliazhko2024mgpt,
  title={mgpt: Few-shot learners go multilingual},
  author={Shliazhko, Oleh and Fenogenova, Alena and Tikhonova, Maria and Kozlova, Anastasia and Mikhailov, Vladislav and Shavrina, Tatiana},
  journal={Transactions of the Association for Computational Linguistics},
  volume={12},
  pages={58--79},
  year={2024},
  publisher={MIT Press One Broadway, 12th Floor, Cambridge, Massachusetts 02142, USA~…}
}

@inproceedings{kondratyuk201975,
  title={75 Languages, 1 Model: Parsing Universal Dependencies Universally},
  author={Kondratyuk, Dan and Straka, Milan},
  booktitle={Proceedings of the 2019 Conference on Empirical Methods in Natural Language Processing and the 9th International Joint Conference on Natural Language Processing (EMNLP-IJCNLP)},
  pages={2779--2795},
  year={2019}
}

@inproceedings{ouyang2021ernie,
  title={ERNIE-M: Enhanced multilingual representation by aligning cross-lingual semantics with monolingual corpora},
  author={Ouyang, Xuan and Wang, Shuohuan and Pang, Chao and Sun, Yu and Tian, Hao and Wu, Hua and Wang, Haifeng},
  booktitle={Proceedings of the 2021 conference on empirical methods in natural language processing},
  pages={27--38},
  year={2021}
}

@article{fan2021beyond,
  title={Beyond english-centric multilingual machine translation},
  author={Fan, Angela and Bhosale, Shruti and Schwenk, Holger and Ma, Zhiyi and El-Kishky, Ahmed and Goyal, Siddharth and Baines, Mandeep and Celebi, Onur and Wenzek, Guillaume and Chaudhary, Vishrav and others},
  journal={Journal of Machine Learning Research},
  volume={22},
  number={107},
  pages={1--48},
  year={2021}
}

@article{campbell2008ethnologue,
  title={Ethnologue: Languages of the world},
  author={Campbell, Lyle and Grondona, Ver{\'o}nica},
  journal={Language},
  volume={84},
  number={3},
  pages={636--641},
  year={2008},
  publisher={Linguistic Society of America}
}

@inproceedings{ebrahimi2021adapt,
  title={How to adapt your pretrained multilingual model to 1600 languages},
  author={Ebrahimi, Abteen and von der Wense, Katharina},
  booktitle={Proceedings of the 59th annual meeting of the association for computational linguistics and the 11th international joint conference on natural language processing (volume 1: Long papers)},
  pages={4555--4567},
  year={2021}
}

@inproceedings{wang2022expanding,
  title={Expanding Pretrained Models to Thousands More Languages via Lexicon-based Adaptation},
  author={Wang, Xinyi and Ruder, Sebastian and Neubig, Graham},
  booktitle={Proceedings of the 60th Annual Meeting of the Association for Computational Linguistics (Volume 1: Long Papers)},
  year={2022}
}

@article{dobler2023focus,
  title={FOCUS: Effective embedding initialization for monolingual specialization of multilingual models},
  author={Dobler, Konstantin and De Melo, Gerard},
  journal={arXiv preprint arXiv:2305.14481},
  year={2023}
}

@inproceedings{wang2020extending,
  title={Extending multilingual BERT to low-resource languages},
  author={Wang, Zihan and Karthikeyan, K and Mayhew, Stephen and Roth, Dan},
  booktitle={Findings of the Association for Computational Linguistics: EMNLP 2020},
  pages={2649--2656},
  year={2020}
}

@inproceedings{mccarthy2020johns,
  title={The Johns Hopkins University Bible corpus: 1600+ tongues for typological exploration},
  author={McCarthy, Arya D and Wicks, Rachel and Lewis, Dylan and Mueller, Aaron and Wu, Winston and Adams, Oliver and Nicolai, Garrett and Post, Matt and Yarowsky, David},
  booktitle={Proceedings of the Twelfth Language Resources and Evaluation Conference},
  pages={2884--2892},
  year={2020}
}

@article{rebuffi2017learning,
  title={Learning multiple visual domains with residual adapters},
  author={Rebuffi, Sylvestre-Alvise and Bilen, Hakan and Vedaldi, Andrea},
  journal={Advances in neural information processing systems},
  volume={30},
  year={2017}
}

@inproceedings{abdaoui2020load,
  title={Load What You Need: Smaller Versions of Mutililingual BERT},
  author={Abdaoui, Amine and Pradel, Camille and Sigel, Gr{\'e}goire},
  booktitle={Proceedings of SustaiNLP: Workshop on Simple and Efficient Natural Language Processing},
  pages={119--123},
  year={2020}
}

@inproceedings{minixhofer2022wechsel,
  title={WECHSEL: Effective initialization of subword embeddings for cross-lingual transfer of monolingual language models},
  author={Minixhofer, Benjamin and Paischer, Fabian and Rekabsaz, Navid},
  booktitle={Proceedings of the 2022 Conference of the North American Chapter of the Association for Computational Linguistics: Human Language Technologies},
  pages={3992--4006},
  year={2022}
}

@article{ruder2019survey,
  title={A survey of cross-lingual word embedding models},
  author={Ruder, Sebastian and Vuli{\'c}, Ivan and S{\o}gaard, Anders},
  journal={Journal of Artificial Intelligence Research},
  volume={65},
  pages={569--631},
  year={2019}
}

@article{tran2019english,
  title={From English to foreign languages: transferring pre-trained language models},
  author={Tran, Ke},
  year={2019}
}

@inproceedings{vernikos2021subword,
  title={Subword Mapping and Anchoring across Languages},
  author={Vernikos, Giorgos and Popescu-Belis, Andrei},
  booktitle={Findings of the Association for Computational Linguistics: EMNLP 2021},
  pages={2633--2647},
  year={2021}
}

@inproceedings{de2021good,
  title={As good as new. how to successfully recycle English GPT-2 to make models for other languages},
  author={de Vries, Wietse and Nissim, Malvina},
  booktitle={Findings of the association for computational linguistics: ACL-IJCNLP 2021},
  pages={836--846},
  year={2021}
}

@inproceedings{wu2020all,
  title={Are all languages created equal in multilingual BERT?},
  author={Wu, Shijie and Dredze, Mark},
  booktitle={5th Workshop on Representation Learning for NLP, RepL4NLP 2020 at the 58th Annual Meeting of the Association for Computational Linguistics, ACL 2020},
  pages={120--130},
  year={2020},
  organization={Association for Computational Linguistics (ACL)}
}

@inproceedings{wang2019improving,
  title={Improving Pre-Trained Multilingual Model with Vocabulary Expansion},
  author={Wang, Hai and Yu, Dian and Sun, Kai and Chen, Jianshu and Yu, Dong},
  booktitle={Proceedings of the 23rd Conference on Computational Natural Language Learning (CoNLL)},
  pages={316--327},
  year={2019}
}

@inproceedings{strassel2016lorelei,
  title={LORELEI language packs: Data, tools, and resources for technology development in low resource languages},
  author={Strassel, Stephanie and Tracey, Jennifer},
  booktitle={Proceedings of the Tenth International Conference on Language Resources and Evaluation (LREC'16)},
  pages={3273--3280},
  year={2016}
}

@article{ammarmassively,
  title={Massively Multilingual Word Embeddings},
  author={Ammar, Waleed and Mulcaire, George and Tsvetkov, Yulia and Lample, Guillaume and Dyer, Chris and Smith, Noah A}
}

@article{jawanpuria2019learning,
  title={Learning multilingual word embeddings in latent metric space: a geometric approach},
  author={Jawanpuria, Pratik and Balgovind, Arjun and Kunchukuttan, Anoop and Mishra, Bamdev},
  journal={Transactions of the Association for Computational Linguistics},
  volume={7},
  pages={107--120},
  year={2019},
  publisher={MIT Press One Rogers Street, Cambridge, MA 02142-1209, USA journals-info~…}
}

@article{kementchedjhievao2018generalizing,
  title={Generalizing Procrustes Analysis for Better Bilingual Dictionary Induction},
  author={KementchedjhievaO, Yova and Ruder, Sebastian and Cotterell, Ryan and S{\o}gaardO, Anders},
  journal={CoNLL 2018},
  pages={211},
  year={2018}
}

@article{bojanowski2017enriching,
  title={Enriching word vectors with subword information},
  author={Bojanowski, Piotr and Grave, Edouard and Joulin, Armand and Mikolov, Tomas},
  journal={Transactions of the association for computational linguistics},
  volume={5},
  pages={135--146},
  year={2017},
  publisher={MIT Press One Rogers Street, Cambridge, MA 02142-1209, USA journals-info~…}
}

@article{schonemann1966generalized,
  title={A generalized solution of the orthogonal procrustes problem},
  author={Sch{\"o}nemann, Peter H},
  journal={Psychometrika},
  volume={31},
  number={1},
  pages={1--10},
  year={1966},
  publisher={Springer-Verlag}
}

@inproceedings{lample2018word,
  title={Word translation without parallel data},
  author={Lample, Guillaume and Conneau, Alexis and Ranzato, Marc'Aurelio and Denoyer, Ludovic and J{\'e}gou, Herv{\'e}},
  booktitle={International conference on learning representations},
  year={2018}
}

@inproceedings{vulic2020all,
  title={Are All Good Word Vector Spaces Isomorphic?},
  author={Vuli{\'c}, Ivan and Ruder, Sebastian and S{\o}gaard, Anders},
  booktitle={Proceedings of the 2020 Conference on Empirical Methods in Natural Language Processing (EMNLP)},
  pages={3178--3192},
  year={2020}
}

@inproceedings{siddhant2020xtreme,
  title={Xtreme: A massively multilingual multi-task benchmark for evaluating cross-lingual generalization},
  author={Siddhant, Aditya and Hu, Junjie and Johnson, Melvin and Firat, Orhan and Ruder, Sebastian},
  booktitle={Proceedings of the International Conference on Machine Learning},
  volume={2020},
  pages={4411--4421},
  year={2020}
}

\end{document}